\documentclass[lettersize,journal]{IEEEtran}
\usepackage{cite}
\usepackage{amsmath,amssymb,amsfonts}
\usepackage{graphicx}
\usepackage{textcomp}
\usepackage{xcolor}
\usepackage{color,array}
\usepackage{graphicx}
\usepackage{svg}
\usepackage{cite}
\usepackage{amsmath,amssymb,amsfonts}
\usepackage{textcomp}
\usepackage{xcolor}
\usepackage{multirow}
\usepackage{booktabs}
\usepackage{siunitx}
\usepackage{url}
\usepackage{newfloat}
\usepackage{xcolor}
\usepackage{listings}
\usepackage{hyperref}
\usepackage{amsthm}
\usepackage{enumitem}
\usepackage{subcaption}

\usepackage{algorithm}
\usepackage{algorithmic}

\newtheorem{definition}{Definition}

\usepackage{multirow}
\usepackage{siunitx}
\begin{document}
\title{Efficient Traffic State Prediction With Dynamic \\ Joint Spatio-Temporal Relation Inference}
\author{
Zhifeng~Hao,~\IEEEmembership{Senior Member,~IEEE,} 
Kai~Hu,
Juncai~Zhang$^*$,
Zhidan~Zhao,
and~Zhengming~Chen
\thanks{Zhifeng Hao, Kai Hu, and Zhengming Chen are with the School of Mathematics and Computer Science, Shantou University, Shantou, China (e-mail: haozhifeng@stu.edu.cn; 23khu@stu.edu.cn; zhengmingchen@stu.edu.cn).}%
\thanks{Juncai Zhang is with the School of Computer Science and Technology, Tongji University, Shanghai, China (e-mail: zhangjuncai.chn@gmail.com).}
\thanks{Zhidan Zhao is with the School of Cyberspace Security, Hainan University, Haikou, China (e-mail: zhidan.zhao@hainanu.edu.cn).}%
\thanks{$^*$Corresponding authors. Emails: zhangjuncai.chn@gmail.com}
}


\markboth{Preprint}{Hao \MakeLowercase{\textit{et al.}}: Efficient Traffic Prediction With Dynamic Joint Spatio-Temporal Relation Inference}

\maketitle

\begin{abstract}
Traffic prediction is difficult due to the complex interplay of temporal evolution, spatial interactions, and delayed spatio-temporal propagation over road networks. Existing methods either model spatial and temporal dependencies separately or employ unified spatio-temporal structures, but they often insufficiently characterize how neighboring sensors at historical timestamps influence a target node, while complex joint models may incur high computation. This paper proposes STEI-PCN, an efficient pure convolutional network based on spatio-temporal encoding and relation inference. It first builds a local causal joint spatio-temporal graph to restrict candidate interactions, then uses absolute position and relative distance encodings to infer dynamic edge weights. A single-layer graph convolution with a position-aware gated activation unit captures local joint dependencies, and temporal dilated causal convolutions complement long-range temporal patterns. A multi-view prediction module fuses raw, local propagation, and long-range temporal representations for direct multi-step forecasting. Experiments on PeMS03, PeMS04, PeMS07, PeMS08, and PeMS-Bay under multiple horizons show that STEI-PCN achieves competitive accuracy with moderate parameters and low training and inference costs. Ablation and fluctuation analyses further verify the contributions of the main components and empirically analyze the effects of the training-stage constraints under sharp speed changes. Our code is available at a GitHub link~\url{https://github.com/Jessez2/STEI-PCN}.
\end{abstract}

\begin{IEEEkeywords}
Traffic prediction, dynamic spatio-temporal relation inference, joint spatio-temporal dependency modeling, graph convolutional networks, temporal convolutional networks
\end{IEEEkeywords}

\section{Introduction}

\IEEEPARstart{T}{raffic} prediction aims to forecast future traffic states by exploiting historical traffic observations and the underlying traffic network structure. However, accurate traffic prediction remains challenging due to the intrinsic complexity and nonlinearity of traffic networks, as well as the complex dependencies embedded in traffic data~\cite{zhang2022complex}.

\begin{figure}[!t]
    \centering
    \includegraphics[width=0.8\linewidth]{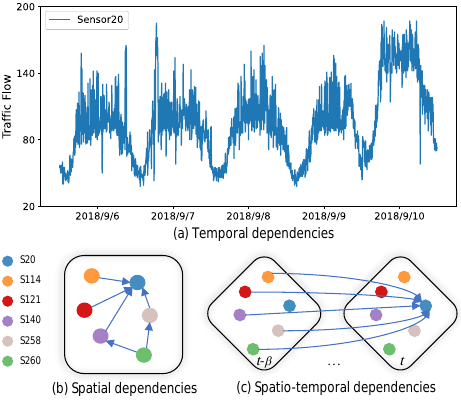}
    \caption{Illustration of temporal, spatial, and spatio-temporal dependencies in traffic data.}
    \label{fig1}
\end{figure}

Traffic networks naturally exhibit a spatio-temporal graph structure~\cite{jin2023spatio}, in which traffic states are coupled across temporal, spatial, and spatio-temporal dimensions. From the temporal perspective, the traffic state of each node usually follows strong periodic patterns. As shown in Fig.~\ref{fig1} (a), the traffic flow collected by sensors presents clear daily regularity, such as the recurrent patterns during morning and evening peak hours. Nevertheless, short-term unexpected events, such as traffic accidents, may lead to abrupt traffic fluctuations, which further increase the difficulty of long-horizon prediction.

From the spatial perspective, the traffic state of a node is not only determined by its own location, but also closely related to the states of its upstream and downstream neighboring nodes. As illustrated in Fig.~\ref{fig1} (b), the traffic state of sensor $S20$ can be affected by directly connected neighbors, and may also be indirectly influenced by more distant nodes through the road network topology.

More importantly, temporal and spatial dependencies are often coupled in real traffic networks. The current traffic state of a node may be affected by the historical states of its neighboring nodes with certain temporal delays. As shown in Fig.~\ref{fig1} (c), the traffic state of sensor $S20$ is potentially influenced not only by neighboring nodes at time $t-1$, but also by those at earlier time steps up to $t-\beta$. Therefore, effectively capturing and integrating temporal, spatial, and spatio-temporal dependencies is essential for building accurate traffic prediction models.

To address the above issues, we propose STEI-PCN, a pure convolutional network based on spatio-temporal encoding and relation inference for both short- and long-horizon traffic prediction. STEI-PCN employs a single-layer graph convolution over the local causal spatio-temporal graph $A_{STP}$~\cite{zhao2022spatial} to capture local joint spatio-temporal propagation, and uses a three-layer temporal dilated causal convolution network (TDCN)~\cite{bai2018empirical} to model long-range temporal dependencies. In addition, a multi-view collaborative prediction module (MVC) is designed to integrate raw, local joint spatio-temporal, and long-range temporal representations for final prediction. 

\noindent \textbf{Contribution.}~The main contributions of this work are summarized as follows:

\begin{itemize}
    \item We propose STEI-PCN, an efficient pure convolutional network for traffic prediction. STEI-PCN separates local joint spatio-temporal propagation modeling from long-range temporal dependency modeling by combining local spatio-temporal graph convolution with temporal dilated causal convolution, thereby maintaining high computational parallelism.

    \item We design a lightweight spatio-temporal encoding and relation inference component, named STEI, to infer dynamic edge weights within the local causal spatio-temporal graph from spatio-temporal position and distance encodings. Compared with the inference function in STPRI~\cite{zhao2022spatial}, STEI reduces the number of additional parameters from $12d$ to $6d$ and the computational complexity from $O(d^2)$ to $O(d)$.

    \item We introduce two training-stage auxiliary constraints for speed prediction under sharp speed changes. The traffic propagation consistency (TPC) constraint encourages target-node representations to be consistent with locally propagated neighborhood states, while the traffic variation fidelity (TVF) constraint preserves the first-order variation direction and magnitude of predicted sequences, helping alleviate over-smoothed predictions in abrupt speed-change scenarios.

    \item We conduct experiments on four traffic flow datasets and one traffic speed dataset. The results show that STEI-PCN achieves competitive prediction accuracy with moderate parameter size and low computational cost. Ablation experiments validate the contributions of STEI, local spatio-temporal graph convolution, TDCN, and multi-view prediction.
\end{itemize}
\noindent \textbf{Outline.} Section~\ref{sec:related_work} reviews related studies on traffic prediction. Section~\ref{sec:preliminaries} introduces the basic definitions and formulates the prediction problem. Section~\ref{sec:method} presents the proposed STEI-PCN, including dynamic joint spatio-temporal relation inference, training objectives, and model properties. Section~\ref{sec:experiments} reports the experimental results and analyses. Section~\ref{sec:conclusion} concludes this paper and discusses future work.

\section{Related Work}
\label{sec:related_work}

This section reviews studies closely related to the proposed model, including traffic prediction methods based on spatio-temporal dependency modeling and those further considering joint spatio-temporal dependencies.

\noindent \textbf{Traffic Prediction With Spatio-Temporal Dependencies.}
Traffic forecasting has been extensively studied in recent years, and several surveys have summarized the development of graph-based and spatio-temporal forecasting models~\cite{jiang2022graph,jin2023spatio,peng2024overview}. Early deep learning-based methods usually adopt separate architectures to model spatial and temporal dependencies with different modules. DCRNN~\cite{li2018dcrnn_traffic} introduces diffusion convolution into recurrent neural networks to model traffic diffusion and temporal dynamics. STGCN~\cite{yu2018spatio} combines graph convolution with temporal convolution, while ASTGCN~\cite{guo2019attention} and GMAN~\cite{zheng2020gman} further incorporate attention mechanisms to enhance spatio-temporal feature extraction. Although these methods achieve promising results, they mainly rely on predefined graph structures, which limit their ability to adapt to time-varying traffic relationships.

To improve relation learning, adaptive and dynamic graph models have been widely studied. GWN~\cite{wu2019graph} and MTGNN~\cite{wu2020connecting} learn adaptive adjacency matrices from node embeddings. AGCRN~\cite{bai2020adaptive} introduces node-adaptive parameters into recurrent graph convolution, and DGCRN~\cite{li2023dynamic} further learns dynamic graph relationships. DSTAGNN~\cite{lan2022dstagnn}, PDFormer~\cite{jiang2023pdformer}, and RGDAN~\cite{fan2024rgdan} also enhance traffic prediction by modeling dynamic spatial or spatio-temporal dependencies. In addition, embedding-based, normalization-based, and pre-training-based methods, such as STID~\cite{shao2022spatial}, STAEformer~\cite{liu2023spatio}, STNorm~\cite{deng2021st}, STEP~\cite{shao2022pre}, and STD-MAE~\cite{gao2023spatial}, have been proposed to improve representation learning. These methods enhance modeling flexibility, but many learned dependencies are still implicit or global, making it difficult to explicitly describe local delayed propagation between neighboring road segments and historical timestamps.

Another line of research extracts temporal and spatial features within unified modules. STSGCN~\cite{song2020spatial} constructs local synchronous spatio-temporal graphs to jointly aggregate information across adjacent timestamps, and STFGCN~\cite{li2022adaptive} builds a unified spatio-temporal fusion graph to capture local and global dependencies. Recent studies have further explored Transformer-enhanced graph models, trend-event decoupling, congestion propagation modeling, cross-scenario adaptation, and non-stationary temporal variation modeling~\cite{zhang2025spatio,chen2025multi,xu2025trend,lin2026dynamic,zhong2026wavegformer,wang2026unlocking,wan2024tcdformer}. These studies show that dynamic dependency modeling and temporal variation modeling remain important, but the influence of neighboring nodes at delayed historical timestamps is still not fully characterized.

\noindent \textbf{Traffic Prediction With Joint Spatio-Temporal Dependencies.}
Joint spatio-temporal dependency modeling further considers the influence of neighboring nodes at historical timestamps, which is more consistent with real traffic evolution. Classical traffic flow theories, including kinematic wave theory, shock wave analysis, and the cell transmission model, have shown that congestion and speed disturbances may propagate along road networks with spatial and temporal delays~\cite{lighthill1955kinematic,richards1956shock,daganzo1994cell}. Recent delay-aware traffic forecasting studies also indicate the importance of delayed spatio-temporal effects~\cite{long2024stdde}. Therefore, an effective model should not only capture temporal and spatial dependencies separately, but also characterize local cross-node and cross-time interactions. This motivates us to decompose the prediction task into identifying local causal spatio-temporal interactions, estimating their dynamic strengths, and integrating them with long-range temporal patterns for multi-step forecasting.

Some methods extract temporal, spatial, and spatio-temporal features with separate components. DST-GraphSAGE~\cite{liu2023graphsage} generates spatio-temporal embeddings from historical observations of spatial neighbors and uses temporal dilated causal convolutions to capture long-range temporal dependencies. SSGCRTN~\cite{yang2024ssgcrtn} combines spatial-specific graph convolution, spatio-temporal interaction modeling, and Transformer-based temporal fusion. Although these methods enhance multi-type dependency modeling, their feature extraction and fusion processes are relatively complex.

Unified joint spatio-temporal modeling methods attempt to capture temporal, spatial, and spatio-temporal dependencies within a single structure. ConSTGAT~\cite{fang2020constgat} uses attention mechanisms to implicitly learn synchronous joint spatio-temporal relationships. In contrast, graph convolution-based methods explicitly encode local joint spatio-temporal relations into graph structures. STPGCN~\cite{zhao2022spatial} constructs a local joint spatio-temporal graph $A_{STP}$ and designs STPRI to infer correlation weights from spatio-temporal position and distance encodings. This explicit structure is consistent with traffic propagation patterns and provides better interpretability.

Nevertheless, STPGCN still relies on a relatively heavy relation inference function and stacks multiple graph convolution layers to enlarge the receptive field, which may increase computational cost and aggravate feature smoothing. Inspired by its explicit joint spatio-temporal modeling idea, we develop a more efficient framework. Specifically, we retain the local causal support of $A_{STP}$, infer dynamic edge weights only within the local spatio-temporal neighborhood, use a single-layer graph convolution to capture local delayed propagation, and employ temporal dilated causal convolution to complement long-range temporal dependency modeling.

\section{Preliminaries and Problem Formulation}
\label{sec:preliminaries}

This section introduces the basic notations used in this paper and formulates the traffic prediction problem.

\subsection{Traffic Network}

\begin{definition}[\textbf{Traffic Graph and Graph Distance}]
A traffic network is represented as a directed or undirected graph $G=(V,E,A)$, where $V=\{v_1,v_2,\ldots,v_N\}$ denotes the set of traffic nodes, $E\subseteq\{(v_i,v_j)\mid v_i,v_j\in V, i\neq j\}$ denotes the set of edges between adjacent nodes, and $A\in\mathbb{R}^{N\times N}$ is the adjacency matrix. In this paper, the adjacency matrix is used to describe the topological support of the traffic network. Specifically, $A_{ij}>0$ if $(v_i,v_j)\in E$, and $A_{ij}=0$ otherwise. For a binary topology, the weights of connected edges can be set to $1$.
The graph distance between nodes $v_i$ and $v_j$ is defined as $d_G(v_i,v_j)=\min\{m \mid \text{there exists a path of length } m \text{ from } v_i \text{ to } v_j\}$.

If $v_i$ and $v_j$ are disconnected, we set $d_G(v_i,v_j)=+\infty$. For a directed graph, $d_G(v_i,v_j)$ denotes the shortest path length along the given edge direction. When an undirected topological support is used, the distance is computed on the undirected version of the graph. Given the spatial interaction order $\alpha$, the $\alpha$-hop neighborhood of node $v_i$ is defined as
\begin{equation}
\mathcal{N}_i^{(\alpha)}=\{v_j\in V \mid d_G(v_i,v_j)\leq \alpha\}.
\end{equation}
\end{definition}

\begin{definition}[\textbf{Traffic State Tensor}]
Assume that the traffic network $G$ generates $F$-dimensional traffic features over $T$ time steps. The traffic state tensor is denoted as
\begin{equation}
X=[X_1,\ldots,X_T]\in\mathbb{R}^{T\times N\times F},
\end{equation}
where $X_t\in\mathbb{R}^{N\times F}$ represents the traffic states of all nodes at time step $t$, and $x_i^t\in\mathbb{R}^{F}$ denotes the traffic state of node $v_i$ at time step $t$. In the experiments of this paper, both traffic flow and traffic speed prediction correspond to the case of $F=1$.
\end{definition}

\begin{definition}[\textbf{Local Causal Joint Spatio-Temporal Graph}]
Given the target time step $t$, the spatial interaction range $\alpha$, and the temporal interaction range $\beta$, the local causal joint spatio-temporal graph is denoted as
\begin{equation}
G_{STP}^t=(V_{STP}^t,E_{STP}^t),
\end{equation}
where
\begin{equation}
V_{STP}^t=\{v_j^\tau \mid v_j\in V,\ \tau\in\{t-\beta,t-\beta+1,\ldots,t\}\},
\end{equation}
and the edge set is defined as
\begin{equation}
E_{STP}^t=
\{(v_j^\tau,v_i^t)\mid v_i,v_j\in V,\ d_G(v_i,v_j)\leq\alpha,\ t-\beta\leq\tau\leq t\}.
\end{equation}
According to this definition, all edges are directed from historical or current time steps $\tau\leq t$ to the target time step $t$. Therefore, the graph satisfies temporal causality. When $t-\beta<1$, the out-of-range historical positions are padded with zeros in implementation.
\end{definition}

\subsection{Problem Formulation}

\begin{definition}[\textbf{Traffic Prediction Function}]
Traffic prediction aims to forecast future traffic states of multiple nodes in a road network based on historical observations. Let $t_0$ denote the last timestamp of the historical input window. Given the traffic graph $G$, the historical input sequence with length $T_h$ is defined as
\begin{equation}
X^H=[X_{t_0-T_h+1},\ldots,X_{t_0}]
\in\mathbb{R}^{T_h\times N\times F}.
\end{equation}
The ground-truth traffic states of the next $T_p$ time steps are denoted as
\begin{equation}
X^P=[X_{t_0+1},\ldots,X_{t_0+T_p}]
\in\mathbb{R}^{T_p\times N\times F}.
\end{equation}
The traffic prediction model is formulated as a prediction operator parameterized by $\Theta$:
\begin{equation}
f_{\Theta}:
\mathbb{R}^{T_h\times N\times F}\times G
\rightarrow
\mathbb{R}^{T_p\times N\times F}.
\end{equation}
The objective of traffic prediction is to learn $f_\Theta$ such that
\begin{equation}
\hat{X}^P=f_\Theta(X^H;G),
\end{equation}
where $\hat{X}^P$ denotes the predicted future traffic states. Here, $T_h$ is the historical input length, $T_p$ is the prediction horizon, and $\Theta$ denotes all trainable parameters of the model.
\end{definition}

For clarity, we distinguish the global timestamp and the within-window index. In the local spatio-temporal graph construction, $t$ denotes a generic absolute timestamp. In the training objective, $\ell=1,\ldots,T_h$ denotes the index inside the historical input window, and $\tau$ denotes a source historical position satisfying $\tau\leq \ell$. The future prediction step is indexed by $k=1,\ldots,T_p$.

\section{Proposed Method}
\label{sec:method}

This section presents the overall architecture, key modules, training objective, and model properties of STEI-PCN.

\subsection{Overall Architecture}

\begin{figure*}[!t]
    \centering
    \includegraphics[width=0.8\linewidth]{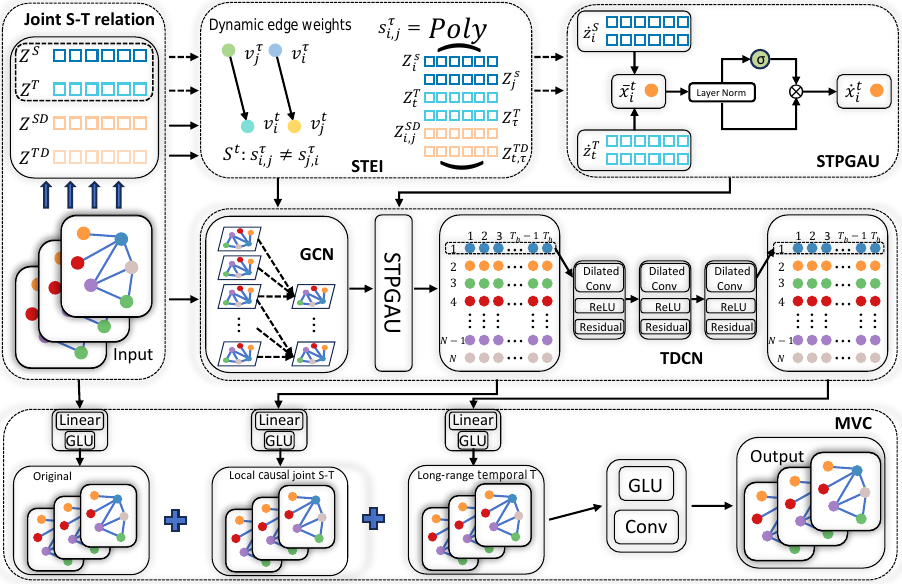}
    \caption{Overall architecture of STEI-PCN.}
    \label{fig:architecture}
\end{figure*}

In this section, we present the overall architecture of STEI-PCN in Fig.~\ref{fig:architecture}. The model consists of four components: joint spatio-temporal relation construction, local joint spatio-temporal dependency modeling, long-range temporal dependency modeling, and multi-view collaborative prediction. Given historical traffic observations, STEI-PCN first maps raw traffic states into hidden representations. It then constructs a local causal joint spatio-temporal graph and infers dynamic edge weights through the spatio-temporal encoding and relation inference component. The inferred graph is used by a single-layer graph convolution to capture local dynamic joint spatio-temporal dependencies, while TDCN is employed to model long-range temporal patterns. Finally, a multi-view collaborative prediction module integrates raw input features, local joint spatio-temporal features, and long-range temporal features to directly generate multi-step predictions.

The design of STEI-PCN follows a division between local propagation modeling and long-range temporal modeling. The local joint spatio-temporal graph restricts direct interactions to causal spatio-temporal neighborhoods, reducing the influence of weakly related distant nodes. TDCN complements this local modeling by enlarging the temporal receptive field. In this way, local propagation patterns and long-range temporal regularities are handled by different modules. When TPC and TVF are used, they act only as training-stage auxiliary constraints and do not modify the forward inference structure or introduce additional online prediction cost.

\subsection{Joint Spatio-Temporal Relation Construction}

\begin{figure}[!t]
    \centering
    \includegraphics[width=0.8\linewidth]{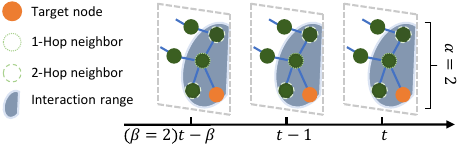}
    \caption{Illustration of local causal joint spatio-temporal graph construction.}
    \label{fig:stp_graph}
\end{figure}

\noindent\textbf{Local Joint Spatio-Temporal Graph.}
To model local joint spatio-temporal dependencies, we construct a local causal joint spatio-temporal graph following the traffic propagation property and the local spatio-temporal graph design in~\cite{zhao2022spatial}. As shown in Fig.~\ref{fig:stp_graph}, the target node at timestamp $t$ is connected to the historical states of its neighboring nodes within a spatial range $\alpha$ and a temporal range $\beta$.

Let $A^{(\alpha)}\in\mathbb{R}^{N\times N}$ denote the $\alpha$-hop spatial adjacency matrix. The local joint spatio-temporal support matrix at timestamp $t$ is defined as
\begin{equation}
M_{STP}^t
=
\big[
A^{(\alpha)}_{t,t}\ \ A^{(\alpha)}_{t,t-1}\ \ \cdots\ \ A^{(\alpha)}_{t,t-\beta}
\big]
\in \mathbb{R}^{N\times(\beta+1)N},
\end{equation}
where $A^{(\alpha)}_{t,\tau}=A^{(\alpha)}$ denotes the spatial support from timestamp $\tau$ to the target timestamp $t$, and $\tau\in\{t-\beta,\ldots,t\}$. Since all edges are directed from historical or current timestamps to the target timestamp, the constructed graph satisfies temporal causality. The elements of $A^{(\alpha)}$ are given by
\begin{equation}
A^{(\alpha)}_{ij}
=
\begin{cases}
1, & \text{if } d_G(v_i,v_j)\leq\alpha,\\
0, & \text{otherwise},
\end{cases}
\end{equation}
where $d_G(v_i,v_j)$ denotes the shortest path distance between nodes $v_i$ and $v_j$ in the traffic graph. The hyperparameters $\alpha$ and $\beta$ control the spatial and temporal interaction ranges, respectively, providing a trade-off between propagation coverage and computational cost.

\noindent\textbf{Spatio-Temporal Encoding.}
To infer dynamic edge weights on the local joint spatio-temporal graph, STEI-PCN uses four types of learnable encodings: absolute spatial coordinate encoding, absolute temporal coordinate encoding, relative spatial distance encoding, and relative temporal distance encoding.

For absolute spatial coordinate encoding, each node $v_i$ is assigned a learnable vector $z_i^S\in\mathbb{R}^d$. For absolute temporal coordinate encoding, each timestamp $t$ is represented by an intra-day index $m\in\{1,2,\ldots,T_{\mathrm{day}}\}$ and a day-of-week index $n\in\{1,2,\ldots,7\}$. The corresponding learnable vectors are denoted as $z_m^D\in\mathbb{R}^d$ and $z_n^W\in\mathbb{R}^d$, and the temporal encoding is obtained by
\begin{equation}
z_t^T=z_m^D+z_n^W.
\end{equation}
For relative spatial distance encoding, each spatial distance $a\in\{0,1,\ldots,\alpha\}$ is associated with a learnable vector $z_a^{SD}\in\mathbb{R}^d$. For relative temporal distance encoding, each temporal distance $b\in\{0,1,\ldots,\beta\}$ is associated with a learnable vector $z_b^{TD}\in\mathbb{R}^d$. These encodings jointly characterize node identity, temporal periodicity, spatial proximity, and temporal delay, which are used to infer dynamic joint spatio-temporal edge weights.

\subsection{Local Joint Spatio-Temporal Dependency Modeling}

\noindent\textbf{Spatio-temporal Encoding and Relation Inference.}
The spatio-temporal encoding and relation inference component, named STEI, infers dynamic interaction strengths for the retained local joint spatio-temporal edges. In Fig.~\ref{fig:architecture}, $\mathrm{Poly}(\cdot)$ is used only as a compact visual notation for the STEI relation inference process. In the actual implementation, the edge weight is computed from six spatio-temporal encoding inputs. For a target node $v_i^t$ and its neighboring node $v_j^\tau$ within the interaction range $(\alpha,\beta)$, where $\tau\in\{t-\beta,\ldots,t\}$, six encoding inputs are defined as
\begin{equation}
\begin{aligned}
z_{ij,1}^{t,\tau} &= z_i^S,        & z_{ij,2}^{t,\tau} &= z_j^S,\\
z_{ij,3}^{t,\tau} &= z_t^T,        & z_{ij,4}^{t,\tau} &= z_\tau^T,\\
z_{ij,5}^{t,\tau} &= z_{d_G(v_i,v_j)}^{SD}, &
z_{ij,6}^{t,\tau} &= z_{t-\tau}^{TD}.
\end{aligned}
\end{equation}
For the $k$-th encoding input, the inference function is defined as
\begin{equation}
\phi_k(z)=\exp(-\|z-\mu_k\|_2),\quad k=1,2,\ldots,6,
\end{equation}
where $\mu_k\in\mathbb{R}^{d}$ is a trainable vector. The dynamic weight of the edge $(v_j^\tau,v_i^t)$ is then computed as
\begin{equation}
s_{ij}^{t,\tau}
=
\mathbf{1}_{\{d_G(v_i,v_j)\leq\alpha,\ t-\beta\leq\tau\leq t\}}
\sum_{k=1}^{6}\phi_k(z_{ij,k}^{t,\tau}),
\end{equation}
where $\mathbf{1}_{\{\cdot\}}$ is the indicator function. Since $0<\phi_k(z)\leq1$, the inferred weight satisfies
\begin{equation}
0\leq s_{ij}^{t,\tau}\leq 6.
\end{equation}
Let $S^t=[s_{ij}^{t,\tau}]_{i,(j,\tau)}\in\mathbb{R}^{N\times(\beta+1)N}$ denote the dynamic edge-weight matrix on the local causal spatio-temporal support. The weighted local spatio-temporal support used for graph convolution is given by
\begin{equation}
\hat{A}_{STP}^{t}=S^t.
\end{equation}

STEI infers edge weights only on the support defined by $M_{STP}^t$, rather than learning a fully connected dynamic graph. This preserves the flexibility of dynamic relation modeling while constraining the candidate interactions by traffic topology and temporal causality. The six inference functions introduce $6d$ additional parameters, excluding the shared encoding tables. Compared with the $12d$ parameters used by the inference function in STPRI~\cite{zhao2022spatial}, STEI reduces the inferring-function parameters by half and decreases the complexity from $O(d^2)$ to $O(d)$.

\noindent\textbf{Local Causal Joint Spatio-Temporal Graph Convolution.}
Given the inferred dynamic edge-weight matrix $\hat{A}_{STP}^t$, a single-layer graph convolution is used to aggregate local causal spatio-temporal neighborhood information. Let $\mathbf{x}_j^\tau\in\mathbb{R}^{F}$ denote the raw traffic observation of node $v_j$ at timestamp $\tau$. It is first transformed into a $C$-dimensional hidden feature by
\begin{equation}
h_j^{\tau}
=
\mathbf{W}^0\mathbf{x}_j^\tau+\mathbf{b}^0,
\end{equation}
where $h_j^\tau\in\mathbb{R}^{C}$ is the encoded feature, and $\mathbf{W}^0\in\mathbb{R}^{C\times F}$ and $\mathbf{b}^0\in\mathbb{R}^{C}$ are trainable parameters.

Let $\widetilde{\mathbf{H}}^H_t
=
\left[
(\mathbf{H}^{H}_{t-\beta})^\top,
\cdots,
(\mathbf{H}^{H}_{t})^\top
\right]^\top
\in \mathbb{R}^{(\beta+1)N\times C}$, where $\mathbf{H}^{H}_{\tau}=[h_1^\tau,\ldots,h_N^\tau]^\top\in\mathbb{R}^{N\times C}$. The aggregation is formulated as
\begin{equation}
\widetilde{\mathbf{X}}^H_t
=
\hat{A}_{STP}^t\widetilde{\mathbf{H}}^H_t
=
S^t\widetilde{\mathbf{H}}^H_t .
\end{equation}
Equivalently, for node $v_i$ at timestamp $t$, the aggregation operator can be written as
\begin{equation}
\mathcal{G}_{STP}(H)_i^t
=
\tilde{x}_i^{t}
=
\sum_{\tau=t-\beta}^{t}
\sum_{v_j\in \mathcal{N}_i^{(\alpha)}}
s_{ij}^{t,\tau}h_j^{\tau}.
\end{equation}
After aggregation, the node feature is updated as
\begin{equation}
\overline{x}_i^{t}
=
\mathbf{W}^1\tilde{x}_i^{t}+\mathbf{b}^1,
\end{equation}
where $\mathbf{W}^1\in\mathbb{R}^{C\times C}$ and $\mathbf{b}^1\in\mathbb{R}^{C}$ are trainable parameters.

This aggregation differs from conventional spatial GCNs because each term contains both a spatial neighbor and a historical timestamp. Thus, the target node receives information from its local causal spatio-temporal neighborhood, enabling the model to describe delayed propagation patterns. In this paper, only one graph convolution layer is used, since the constructed local joint spatio-temporal graph already covers $\alpha$-hop spatial neighbors and $\beta+1$ temporal positions. Stacking more graph convolution layers may enlarge the interaction range implicitly and increase the risk of over-smoothing.

To recalibrate the aggregated representation with node and timestamp identities, we use the spatio-temporal position-aware gated activation unit (STPGAU)~\cite{zhao2022spatial}. Define
\begin{equation}
g_i^t
=
\overline{x}_i^{t}
+
\mathbf{W}^{S}z_i^S
+
\mathbf{W}^{T}z_t^T
+
\mathbf{b}^{g},
\end{equation}
where $\mathbf{W}^{S},\mathbf{W}^{T}\in\mathbb{R}^{C\times d}$ and $\mathbf{b}^{g}\in\mathbb{R}^{C}$ are trainable parameters. The gated output is computed as
\begin{equation}
\dot{x}_i^{t}
=
(\mathbf{W}^4g_i^t+\mathbf{b}^4)
\otimes
\sigma(\mathbf{W}^5g_i^t+\mathbf{b}^5),
\end{equation}
where $\otimes$ denotes element-wise multiplication and $\sigma(\cdot)$ denotes the sigmoid function. STPGAU uses absolute spatial and temporal encodings to guide channel-wise feature selection, helping preserve node-specific and time-specific patterns after neighborhood aggregation.

\subsection{Long-Range Temporal Dependency Modeling}

After local joint spatio-temporal aggregation, TDCN is used to capture long-range temporal dependencies. Compared with recurrent structures, dilated causal convolution supports parallel computation and avoids using future information.

Let $\dot{\mathbf{X}}_i=[\dot{x}_i^{1},\dot{x}_i^{2},\ldots,\dot{x}_i^{T_h}]$ denote the output sequence of node $v_i$ after local joint spatio-temporal modeling. Let $\dot{x}_i^{t,l}$ denote the input feature of node $v_i$ at timestamp $t$ in the $l$-th TDCN layer. With dilation factor $r_l$ and kernel size $K$, the dilated causal convolution is defined as
\begin{equation}
\dot{x}_i^{t,l+1}
=
\operatorname{ReLU}
\left(
\sum_{q=1}^{K}
\mathbf{W}_q^{l}
\dot{x}_i^{t-(q-1)r_l,l}
+
\mathbf{b}^{l}
\right),
\end{equation}
where causal padding is used when $t-(q-1)r_l$ falls outside the input range.

\begin{figure}[!t]
    \centering
    \includegraphics[width=0.8\linewidth]{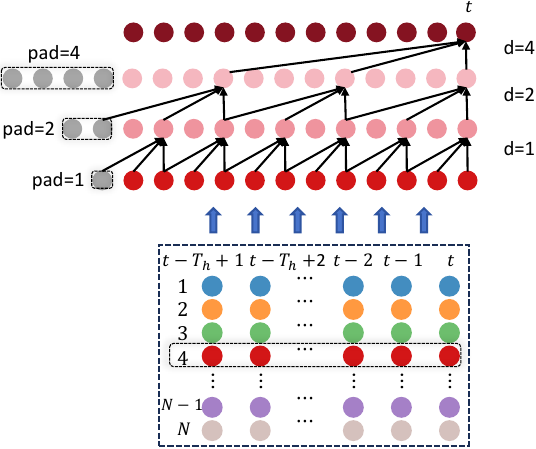}
    \caption{Structure of the temporal dilated causal convolution network.}
    \label{fig:tdcn}
\end{figure}

STEI-PCN stacks three TDCN layers with kernel size $3\times1$ and dilation factors $(1,2,4)$. The labels $d=1,2,4$ in Fig.~\ref{fig:tdcn} denote dilation factors only, while $d$ in the encoding module denotes the encoding dimension. Residual connections~\cite{he2016deep} are adopted in each layer to improve training stability. A two-dimensional convolution layer is further used to adjust the output channel dimension. This design allows the model to enlarge the temporal receptive field with a small number of layers, complementing the local propagation modeling performed by graph convolution.

\subsection{Multi-View Collaborative Prediction}

STEI-PCN uses a multi-view collaborative prediction module to integrate three types of representations: raw input features, local joint spatio-temporal features, and long-range temporal features. For the $m$-th view, let
\begin{equation}
V^{(m)}
=
[\chi^{(m)}_1,\chi^{(m)}_2,\ldots,\chi^{(m)}_N]
\in \mathbb{R}^{T_h\times N\times C}.
\end{equation}
A two-dimensional convolution is first applied to compress the temporal dimension:
\begin{equation}
\widetilde{V}^{(m)}
=
\mathbf{W}^{6}V^{(m)}+\mathbf{b}^{6},
\end{equation}
where $\widetilde{V}^{(m)}\in\mathbb{R}^{C\times N\times 1}$. To suppress redundant feature channels and enhance nonlinear feature selection, we adopt the gated linear unit (GLU), following the gated convolutional design in~\cite{dauphin2017language}. The view-specific GLU output is computed as
\begin{equation}
U^{(m)}
=
(\mathbf{W}^{7}\widetilde{V}^{(m)}+\mathbf{b}^{7})
\otimes
\sigma(\mathbf{W}^{8}\widetilde{V}^{(m)}+\mathbf{b}^{8}).
\end{equation}
The three view-specific outputs are concatenated as
\begin{equation}
U
=
\operatorname{Concat}
\left(
U^{(1)},
U^{(2)},
U^{(3)}
\right).
\end{equation}
Another GLU is then applied to enhance nonlinear feature fusion. Finally, a two-dimensional convolution maps the fused representation to the prediction space and produces
\begin{equation}
\hat{X}^P\in\mathbb{R}^{T_p\times N\times1}.
\end{equation}
Different from recursive prediction strategies, STEI-PCN directly outputs the entire $T_p$-step sequence in one forward pass, which avoids error accumulation and maintains high inference parallelism.

\subsection{Training Objective and Algorithm}
\label{subsec:training_objective}

The training objective first requires the predicted traffic states to be close to the ground truth in terms of point-wise numerical errors. For generality, we denote the basic point-wise loss as
\begin{equation}
\mathcal{L}_{\rho}
=
\frac{1}{NT_p}
\sum_{k=1}^{T_p}
\sum_{i=1}^{N}
\rho\left(
\hat{X}^{P}(k,i)-X^{P}(k,i)
\right),
\end{equation}
where $\rho(\cdot)$ is a point-wise error function. In this paper, we instantiate $\rho(e)=|e|$, leading to the MAE loss
\begin{equation}
\mathcal{L}_{\mathrm{MAE}}
=
\frac{1}{NT_p}
\sum_{k=1}^{T_p}
\sum_{i=1}^{N}
\left|
\hat{X}^{P}(k,i)-X^{P}(k,i)
\right|.
\end{equation}
MAE is adopted in the experiments because it is consistent with the main evaluation metric and is less dominated by a few abrupt traffic variations than squared-error losses. However, point-wise losses mainly measure numerical deviations and do not explicitly regularize local propagation consistency or temporal variation patterns. Therefore, two optional training-stage constraints, TPC and TVF, are introduced.

Let
\begin{equation}
H=\{h_i^\ell\mid i=1,\ldots,N,\ \ell=1,\ldots,T_h\}
\end{equation}
denote the hidden representations after local joint spatio-temporal graph convolution and STPGAU, where $h_i^\ell=\dot{x}_i^\ell$. Here, $\ell$ denotes the index within the historical input window. To avoid the trivial self-connection and invalid padded timestamps, the effective propagation neighborhood of target node $v_i^\ell$ is defined as
\begin{equation}
\begin{split}
\mathcal{N}_{\alpha,\beta}^{-}(v_i^\ell)
=
\{v_j^\tau \mid &\, v_j\in\mathcal{N}_i^{(\alpha)},\
\max(1,\ell-\beta)\leq \tau \leq \ell,\\
& (j,\tau)\neq(i,\ell)\}.
\end{split}
\end{equation}
This definition masks out invalid historical positions in the propagation-state computation. Based on the STEI-inferred edge weight $s_{ij}^{\ell,\tau}$, the normalized propagation weight is
\begin{equation}
\bar{s}_{ij}^{\ell,\tau}
=
\frac{s_{ij}^{\ell,\tau}}
{\sum_{v_k^\rho\in \mathcal{N}_{\alpha,\beta}^{-}(v_i^\ell)}
s_{ik}^{\ell,\rho}+\varepsilon},
\end{equation}
where $\varepsilon$ is a small constant. The local spatio-temporal propagation state of $v_i^\ell$ is computed as
\begin{equation}
\mathcal{P}_{ST}(H)_i^\ell
=
\sum_{v_j^\tau\in \mathcal{N}_{\alpha,\beta}^{-}(v_i^\ell)}
\bar{s}_{ij}^{\ell,\tau}h_j^\tau.
\end{equation}
The traffic propagation consistency constraint is defined as
\begin{equation}
\mathcal{R}_{TPC}
=
\frac{1}{NT_h}
\sum_{\ell=1}^{T_h}
\sum_{i=1}^{N}
\left\|
h_i^\ell-\mathcal{P}_{ST}(H)_i^\ell
\right\|_2^2.
\end{equation}
TPC is a weak representation-level constraint inspired by local traffic propagation. It does not impose a strict traffic-flow physical equation.

To further constrain short-term traffic-state variations, TVF compares the first-order temporal differences of the predicted and ground-truth sequences. The ground-truth variation is defined as
\begin{equation}
\Delta X^P(k,i)
=
\begin{cases}
X^P(1,i)-X^H(T_h,i), & k=1,\\
X^P(k,i)-X^P(k-1,i), & 2\leq k\leq T_p,
\end{cases}
\end{equation}
and the predicted variation is defined as
\begin{equation}
\Delta \hat{X}^P(k,i)
=
\begin{cases}
\hat{X}^P(1,i)-X^H(T_h,i), & k=1,\\
\hat{X}^P(k,i)-\hat{X}^P(k-1,i), & 2\leq k\leq T_p.
\end{cases}
\end{equation}
The ground-truth variation magnitude is
\begin{equation}
a_i^k
=
\left\|
\Delta X^P(k,i)
\right\|_1.
\end{equation}
To assign larger weights to stronger traffic-state variations, the fluctuation weight is defined as
\begin{equation}
\omega_i^k
=
1+
\eta
\frac{a_i^k}
{\varepsilon+
\frac{1}{NT_p}
\sum_{r=1}^{T_p}
\sum_{j=1}^{N}
a_j^r},
\end{equation}
where $\eta\geq0$ controls the additional weight. The TVF constraint is then defined as
\begin{equation}
\mathcal{R}_{TVF}
=
\frac{1}{NT_p}
\sum_{k=1}^{T_p}
\sum_{i=1}^{N}
\omega_i^k
\left\|
\Delta \hat{X}^P(k,i)-\Delta X^P(k,i)
\right\|_1.
\end{equation}
TVF encourages the predicted sequence to preserve the direction and magnitude of traffic-state changes, rather than simply producing smoother predictions.

Combining the basic point-wise loss and the two auxiliary constraints, the overall training objective is
\begin{equation}
\mathcal{L}
=
\mathcal{L}_{\rho}
+
\lambda\mathcal{R}_{TPC}
+
\gamma\mathcal{R}_{TVF},
\end{equation}
where $\lambda$ and $\gamma$ control the strengths of TPC and TVF, respectively. In the current experiments, $\mathcal{L}_{\rho}$ is instantiated as $\mathcal{L}_{\mathrm{MAE}}$. The basic STEI-PCN corresponds to $\lambda=\gamma=0$. STEI-PCN+TPC, STEI-PCN+TVF, and STEI-PCN+TPC+TVF correspond to $(\lambda>0,\gamma=0)$, $(\lambda=0,\gamma>0)$, and $(\lambda>0,\gamma>0)$, respectively. Both TPC and TVF are used only during training and are removed during inference.

Algorithm~\ref{alg:steipcn} summarizes the training and inference procedure of STEI-PCN.

\begin{algorithm}[ht]
\caption{Training and Inference Framework of STEI-PCN}
\label{alg:steipcn}
\textbf{Input}: Traffic graph $G=(V,E,A)$, historical input $X^H$, future label $X^P$, spatial range $\alpha$, temporal range $\beta$, encoding dimension $d$, hidden channel size $C$, and loss weights $\lambda$, $\gamma$, $\eta$.\\
\textbf{Output}: Optimized model parameters $\Theta$, and the predicted future traffic states $\hat{X}^{P}=f_{\Theta}(X^H;G)$ for a given historical input.
\begin{algorithmic}[1]
\STATE Initialize all trainable parameters in STEI, GCN, STPGAU, TDCN, and MVC.
\FOR{each training batch}
    \STATE Construct the local causal joint spatio-temporal support using $G$, $\alpha$, and $\beta$.
    \STATE Infer dynamic edge weights $s_{ij}^{\ell,\tau}$ on the local support with STEI.
    \STATE Generate $\hat{X}^{P}$ through local graph convolution, STPGAU, TDCN, and MVC.
    \STATE Compute the basic point-wise loss $\mathcal{L}_{\rho}$.
    \STATE If $\lambda>0$, compute $\mathcal{R}_{TPC}$ on the local propagation neighborhood excluding the self-edge and invalid padded timestamps.
    \STATE If $\gamma>0$, compute $\mathcal{R}_{TVF}$ using the first-order differences of $\hat{X}^{P}$ and $X^{P}$.
    \STATE Update $\Theta$ by minimizing
    $\mathcal{L}=\mathcal{L}_{\rho}+\lambda\mathcal{R}_{TPC}+\gamma\mathcal{R}_{TVF}$.
\ENDFOR
\STATE During inference, only the forward prediction path is used, and TPC/TVF are not computed.
\STATE \textbf{Return}: $\hat{X}^{P}$.
\end{algorithmic}
\end{algorithm}

\subsection{Model Properties and Complexity}
\label{subsec:model_properties}

This subsection summarizes the main properties of STEI-PCN in terms of temporal causality, sparse relation inference, bounded edge weights, and computational cost.

First, STEI-PCN is temporally causal during inference. Given the historical input
\begin{equation}
X^H=[X_{t_0-T_h+1},\ldots,X_{t_0}],
\end{equation}
the prediction is formulated as
\begin{equation}
\hat{X}^{P}=f_{\Theta}(X^H;G),
\end{equation}
where no future ground-truth states are used as input. The local causal joint spatio-temporal graph only aggregates information from current or historical timestamps; TDCN adopts causal convolution, and MVC fuses only historical representations. Although TVF uses future labels to compute the supervised variation loss during training, it is removed during inference and does not introduce information leakage.

Second, STEI infers relations only on a sparse local support. Let $Q_\alpha=\max_i|\mathcal{N}_i^{(\alpha)}|$ denote the maximum number of $\alpha$-hop neighbors. For each target timestamp, the number of candidate spatio-temporal edges is at most
\begin{equation}
N(\beta+1)Q_\alpha.
\end{equation}
Thus, STEI-PCN avoids fully connected space-time relation inference and controls unnecessary interactions. The local graph convolution captures short-range delayed propagation, while TDCN complements longer-range temporal patterns through stacked dilated causal convolutions.

Third, the inferred edge weights are bounded. Since
\begin{equation}
\phi_k(z)=\exp(-\|z-\mu_k\|_2), \quad k=1,\ldots,6,
\end{equation}
we have $0<\phi_k(z)\leq1$. Therefore, for each retained edge, $0<s_{ij}^{t,\tau}\leq6$, while masked edges outside the local support have zero weights. Moreover, for two encoding input groups $\mathbf{z}=(z_1,\ldots,z_6)$ and $\mathbf{z}'=(z'_1,\ldots,z'_6)$ on the same edge support,
\begin{equation}
|s(\mathbf{z})-s(\mathbf{z}')|
\leq
\sum_{k=1}^{6}\|z_k-z'_k\|_2.
\end{equation}
Thus, under a fixed local support, the inferred weights are stable with respect to small changes in the spatio-temporal encodings.

Finally, the computational cost is mainly determined by local edge search, STEI relation inference, graph aggregation, temporal convolution, and prediction mapping. Let $M$ denote the number of retained local spatio-temporal edges for one target timestamp. Ignoring constant factors, the forward complexity of the basic STEI-PCN is
\begin{equation}
O\left(
\alpha NQ_\alpha
+
dT_hM
+
T_hMC
+
T_hNC^2
+
dT_hNC
+
T_pNF
\right).
\end{equation}
The terms correspond to local neighborhood construction, STEI edge-weight inference, local graph aggregation, feature transformation, encoding-guided gated activation, and multi-step prediction, respectively. TPC and TVF only introduce additional loss computations during training and do not increase the online inference cost.

\section{Experiments}
\label{sec:experiments}

In this section, we evaluate the proposed STEI-PCN on both traffic flow and traffic speed prediction tasks. The experiments are designed to answer the following questions: 1) whether STEI-PCN achieves competitive prediction accuracy compared with representative baselines; 2) whether each component contributes to the final performance; 3) how the proposed TPC and TVF constraints affect speed prediction under sharp speed changes; and 4) whether STEI-PCN maintains reasonable computational efficiency.

\begin{table}[ht]
\centering
\scriptsize
\caption{Statistics of the experimental datasets.}
\label{tab:datasets}
\begin{tabular}{lccccc}
\toprule
Dataset & Nodes & Time Steps & Time Span & Missing Ratio & Type \\
\midrule
PeMS03 & 358 & 26208 & 09/2018--11/2018 & 0.672\% & Flow \\
PeMS04 & 307 & 16992 & 01/2018--02/2018 & 3.182\% & Flow \\
PeMS07 & 883 & 28224 & 05/2017--08/2017 & 0.452\% & Flow \\
PeMS08 & 170 & 17856 & 07/2016--08/2016 & 0.696\% & Flow \\
PeMS-Bay & 325 & 52116 & 01/2017--05/2017 & 0.003\% & Speed \\
\bottomrule
\end{tabular}

\end{table}

\begin{table*}[!]
\centering
\scriptsize
\caption{Performance comparison on PeMS03, PeMS04, PeMS07, and PeMS08 with prediction horizon $T_p=6$. Bold indicates the best result and underline indicates the second-best result.}
\label{tab:flow_results}
\begin{tabular}{lcccccccccccc}
\toprule
\multirow{2}{*}{Model} & \multicolumn{3}{c}{PeMS03} & \multicolumn{3}{c}{PeMS04} & \multicolumn{3}{c}{PeMS07} & \multicolumn{3}{c}{PeMS08} \\
\cmidrule(r){2-4} \cmidrule(r){5-7} \cmidrule(r){8-10} \cmidrule(r){11-13}
 & MAE & RMSE & MAPE & MAE & RMSE & MAPE & MAE & RMSE & MAPE & MAE & RMSE & MAPE \\
\midrule
DCRNN (2017) & 15.54 & 27.18 & 15.62\% & 19.63 & 31.26 & 13.59\% & 21.16 & 34.14 & 9.02\% & 15.22 & 24.17 & 10.21\% \\
STGCN (2017) & 15.83 & 27.51 & 16.13\% & 19.57 & 31.38 & 13.44\% & 21.74 & 35.27 & 9.24\% & 16.08 & 25.39 & 10.60\% \\
ASTGCN (2019) & 17.69 & 29.66 & 19.40\% & 22.93 & 35.22 & 16.56\% & 28.05 & 42.57 & 13.92\% & 18.61 & 28.16 & 13.08\% \\
GWN (2019) & 14.59 & 25.24 & 15.52\% & 18.53 & 29.92 & 12.89\% & 20.47 & 33.47 & 8.61\% & 14.40 & 23.39 & 9.21\% \\
AGCRN (2020) & 15.24 & 26.65 & 15.89\% & 19.38 & 31.25 & 13.40\% & 20.57 & 34.40 & 8.74\% & 15.32 & 24.41 & 10.03\% \\
GMAN (2020) & 16.87 & 27.92 & 18.23\% & 19.14 & 31.60 & 13.19\% & 20.97 & 34.10 & 9.05\% & 15.31 & 24.92 & 10.13\% \\
MTGNN (2020) & 14.85 & 25.23 & 14.55\% & 19.17 & 31.70 & 13.37\% & 20.89 & 34.06 & 9.00\% & 15.18 & 24.24 & 10.20\% \\
STSGCN (2020) & 17.48 & 29.21 & 16.78\% & 21.19 & 33.65 & 13.90\% & 24.26 & 39.03 & 10.21\% & 17.13 & 26.80 & 10.96\% \\
STNorm (2021) & 15.32 & 25.93 & 14.37\% & 18.96 & 30.98 & 12.69\% & 20.50 & 34.66 & 8.75\% & 15.41 & 24.77 & 9.76\% \\
Z-GCNETs (2021) & 16.64 & 28.15 & 16.39\% & 19.50 & 31.61 & 12.78\% & 21.77 & 35.17 & 9.25\% & 15.76 & 25.11 & 10.01\% \\
DSTAGNN (2022) & 15.57 & 27.21 & 14.68\% & 19.30 & 31.46 & 12.70\% & 21.42 & 34.51 & 9.01\% & 15.67 & 24.77 & 9.94\% \\
STEP (2022) & 14.22 & 24.55 & 14.42\% & 18.20 & 29.71 & 12.48\% & 19.32 & 32.19 & 8.12\% & 14.00 & 23.41 & 9.50\% \\
STFGCN (2022) & 16.77 & 28.34 & 16.30\% & 19.83 & 31.88 & 13.02\% & 22.07 & 35.80 & 9.21\% & 16.64 & 26.22 & 10.60\% \\
STID (2022) & 15.33 & 27.40 & 16.40\% & 18.29 & 29.86 & 12.46\% & 19.59 & 32.90 & 8.30\% & 14.21 & 23.57 & 9.24\% \\
STPGCN (2022) & 14.99 & 24.83 & 14.97\% & 18.46 & 30.15 & 12.01\% & 19.70 & 32.99 & 8.19\% & 13.81 & 23.58 & 9.06\% \\
DGCRN (2023) & 14.60 & 26.20 & 14.87\% & 18.84 & 30.48 & 12.92\% & 20.04 & 32.86 & 8.63\% & 14.77 & 23.81 & 9.77\% \\
PDFormer (2023) & 14.94 & 25.39 & 15.82\% & 18.32 & 29.97 & 12.10\% & 19.83 & 32.87 & 8.53\% & 13.58 & 23.51 & 9.05\% \\
STAEformer (2023) & 15.35 & 27.55 & 15.18\% & 18.22 & 30.18 & 11.98\% & 19.14 & 32.60 & 8.01\% & 13.46 & 23.25 & 8.88\% \\
STWave (2023) & 14.92 & 26.70 & 15.53\% & 18.68 & 30.62 & 12.62\% & 19.48 & 33.32 & 8.16\% & 13.69 & 23.47 & 9.40\% \\
STD-MAE (2024) & \underline{13.80} & \underline{24.43} & \textbf{13.96\%} & \textbf{17.80} & \underline{29.25} & 11.97\% & \underline{18.65} & \underline{31.44} & \underline{7.84\%} & 13.44 & \underline{22.47} & 8.76\% \\
HTVGNN (2024) & 14.30 & 24.59 & 14.69\% & 18.01 & 29.81 & \underline{11.89\%} & 19.50 & 32.65 & 8.15\% & 13.28 & 22.83 & \underline{8.65\%} \\
SSGCRTN (2024) & 15.18 & 26.52 & 14.66\% & 19.28 & 31.17 & 12.68\% & 20.71 & 34.22 & 8.69\% & 15.18 & 24.32 & 9.59\% \\
DTRformer (2025) & 14.50 & 25.45 & 14.94\% & \underline{18.00} & 29.58 & 12.30\% & 18.99 & 32.23 & 7.93\% & \underline{13.17} & 22.85 & 8.66\% \\
\textbf{STEI-PCN~(Ours)} & \textbf{13.78} & \textbf{22.17} & \underline{14.26\%} & \textbf{17.80} & \textbf{29.16} & \textbf{11.74\%} & \textbf{18.33} & \textbf{30.66} & \textbf{7.67\%} & \textbf{13.03} & \textbf{21.87} & \textbf{8.50\%} \\
\bottomrule
\end{tabular}
\end{table*}

\subsection{Experimental Setup}

\subsubsection{Datasets}

We conduct experiments on four traffic flow datasets, i.e., PeMS03, PeMS04, PeMS07, and PeMS08~\cite{song2020spatial}, and one traffic speed dataset, i.e., PeMS-Bay~\cite{li2018dcrnn_traffic}. These datasets cover different road-network scales, time spans, missing ratios, and traffic-state types, enabling a comprehensive evaluation of the proposed model.

As shown in Table~\ref{tab:datasets}, the five datasets differ significantly in network scale and traffic-state type. PeMS07 contains the largest number of nodes and is used to examine the scalability of the model on large road networks. PeMS08 is relatively small but still contains typical spatio-temporal traffic fluctuations. PeMS-Bay records traffic speed and is used to further evaluate the model under sharp speed-change scenarios.

\subsubsection{Baselines}

We compare STEI-PCN with representative traffic prediction models, including fixed-graph methods, adaptive and dynamic graph methods, synchronous spatio-temporal graph models, Transformer-based models, pre-training methods, and trend-aware models. Specifically, the compared baselines include DCRNN~\cite{li2018dcrnn_traffic}, STGCN~\cite{yu2018spatio}, ASTGCN~\cite{guo2019attention}, GWN~\cite{wu2019graph}, AGCRN~\cite{bai2020adaptive}, GMAN~\cite{zheng2020gman}, MTGNN~\cite{wu2020connecting}, STSGCN~\cite{song2020spatial}, STNorm~\cite{deng2021st}, Z-GCNETs~\cite{chen2021z}, DSTAGNN~\cite{lan2022dstagnn}, STEP~\cite{shao2022pre}, STFGCN~\cite{li2022adaptive}, STID~\cite{shao2022spatial}, STPGCN~\cite{zhao2022spatial}, DGCRN~\cite{li2023dynamic}, PDFormer~\cite{jiang2023pdformer}, STAEformer~\cite{liu2023spatio}, STWave~\cite{fang2023spatio}, STD-MAE~\cite{gao2023spatial}, HTVGNN~\cite{dai2024novel}, RGDAN~\cite{fan2024rgdan}, SSGCRTN~\cite{yang2024ssgcrtn}, DTRformer~\cite{chen2025dynamic}, and T-Graphormer~\cite{bai2025t}.

These baselines cover a wide range of modeling paradigms. DCRNN, STGCN, and ASTGCN mainly rely on predefined graph structures. GWN, AGCRN, MTGNN, DSTAGNN, DGCRN, HTVGNN, and RGDAN emphasize adaptive or dynamic graph learning. STSGCN, Z-GCNETs, STFGCN, and STPGCN are closely related to synchronous or joint spatio-temporal graph modeling. GMAN, PDFormer, STAEformer, SSGCRTN, and T-Graphormer use attention or Transformer structures to enhance long-range dependency modeling. STEP and STD-MAE learn transferable spatio-temporal representations through pre-training, while STNorm, STWave, and DTRformer focus on normalization, trend decomposition, or non-stationary traffic patterns.

\subsubsection{Implementation Details}

All experiments are conducted on a server equipped with an NVIDIA RTX 4090 GPU with 24GB memory, a 16-core Intel Xeon Gold 6430 CPU, and 120GB RAM. We use $T_h=12$ consecutive historical time steps as input and predict the next $T_p$ time steps, where $T_p\in\{3,6,12\}$. The model is trained using the Adam optimizer with a learning rate of $0.002$.

The main hyperparameters are set as $\alpha=4$, $\beta=2$, $d=6$, $C=64$, kernel size $=3$, and dilation factors $(1,2,4)$. For PeMS03 and PeMS-Bay, the maximum number of training epochs is set to $60$, and the batch size is set to $32$. For PeMS04, PeMS07, and PeMS08, the maximum number of epochs is set to $200$, with batch sizes of $32$, $18$, and $32$, respectively. The smaller batch size on PeMS07 is used due to its larger network scale. PeMS-Bay is split into training, validation, and test sets with a ratio of $7:1:2$, while the other datasets are split with a ratio of $6:2:2$~\cite{song2020spatial}. Unless otherwise specified, the main experiments are repeated three times with different random seeds, and the averaged results are reported. The checkpoint with the best validation MAE is selected for test evaluation.

\subsubsection{Evaluation Metrics}

We use three widely adopted metrics, namely mean absolute error (MAE), root mean square error (RMSE), and mean absolute percentage error (MAPE). For a prediction step $p$, they are defined as
\begin{equation}
\begin{aligned}
\mathrm{MAE} &= \frac{1}{N}\sum_{i=1}^{N}\left|\hat{x}_i^p-x_i^p\right|,\\
\mathrm{RMSE} &= \sqrt{\frac{1}{N}\sum_{i=1}^{N}(\hat{x}_i^p-x_i^p)^2},\\
\mathrm{MAPE} &= \frac{100\%}{N}\sum_{i=1}^{N}\left|\frac{\hat{x}_i^p-x_i^p}{x_i^p}\right|,
\end{aligned}
\end{equation}
where $x_i^p$ and $\hat{x}_i^p$ denote the ground-truth and predicted traffic states of node $v_i$ at prediction step $p$, respectively. When reporting overall results over multiple prediction steps, the metrics are averaged over all prediction steps.

To further evaluate the ability of the model to characterize short-term speed variations, we additionally report direction accuracy (DirAcc) and variation MAE (VarMAE) on PeMS-Bay. Let $\Delta X^P(k,i)$ and $\Delta \hat{X}^P(k,i)$ denote the first-order variations of the ground-truth and predicted future speed sequences, respectively. DirAcc and VarMAE are defined as
\begin{equation}
\begin{split}
\mathrm{DirAcc}
&=
\frac{100\%}{NT_p}
\sum_{k=1}^{T_p}\sum_{i=1}^{N}
\mathbf{1}_{\{\operatorname{sign}(\Delta \hat{X}^P(k,i))=\operatorname{sign}(\Delta X^P(k,i))\}},\\
\mathrm{VarMAE}
&=
\frac{1}{NT_p}
\sum_{k=1}^{T_p}\sum_{i=1}^{N}
\left|\Delta \hat{X}^P(k,i)-\Delta X^P(k,i)\right|.
\end{split}
\end{equation}
In our work, DirAcc measures whether the model correctly predicts the increasing, decreasing, or unchanged direction of speed states, while VarMAE measures the error in speed variation magnitude.

\begin{table*}[ht]
\centering
\scriptsize
\caption{Performance comparison on PeMS-Bay under different prediction horizons. Bold indicates the best result, and underline indicates the second-best result.}
\label{tab:speed_results}
\begin{tabular}{lccccccccc}
\toprule
\multirow{3}{*}{Model} & \multicolumn{3}{c}{$T_p=3$} & \multicolumn{3}{c}{$T_p=6$} & \multicolumn{3}{c}{$T_p=12$} \\
\cmidrule(r){2-4} \cmidrule(r){5-7} \cmidrule(r){8-10}
 & MAE & RMSE & MAPE & MAE & RMSE & MAPE & MAE & RMSE & MAPE \\
\midrule
STEP (2022) & 1.26 & 2.73 & 2.59\% & 1.55 & 3.58 & 3.43\% & 1.79 & \underline{4.20} & 4.18\% \\
STID (2022) & 1.30 & 2.81 & 2.73\% & 1.62 & 3.72 & 3.68\% & 1.89 & 4.40 & 4.47\% \\
PDFormer (2023) & 1.32 & 2.83 & 2.78\% & 1.64 & 3.79 & 3.71\% & 1.91 & 4.43 & 4.51\% \\
STAEformer (2023) & 1.31 & 2.78 & 2.76\% & 1.62 & 3.68 & 3.62\% & 1.88 & 4.34 & 4.41\% \\
STD-MAE (2024) & \underline{1.23} & 2.62 & \underline{2.56\%} & 1.53 & 3.53 & 3.42\% & \underline{1.77} & \underline{4.20} & \underline{4.17\%} \\
RGDAN (2024) & 1.31 & 2.79 & 2.77\% & 1.56 & 3.55 & 3.47\% & 1.82 & \underline{4.20} & 4.28\% \\
T-Graphormer (2025) & 1.31 & \underline{2.55} & 2.71\% & \underline{1.52} & \textbf{3.14} & \underline{3.23\%} & \textbf{1.76} & \textbf{3.78} & \textbf{3.91\%} \\
\textbf{STEI-PCN~(Ours)} & \textbf{1.17} & \textbf{2.46} & \textbf{2.33\%} & \textbf{1.47} & \underline{3.35} & \textbf{3.12\%} & 1.88 & 4.42 & 4.28\% \\
\bottomrule
\end{tabular}
\end{table*}

\begin{figure*}[ht]
    \centering
    \includegraphics[width=1\linewidth]{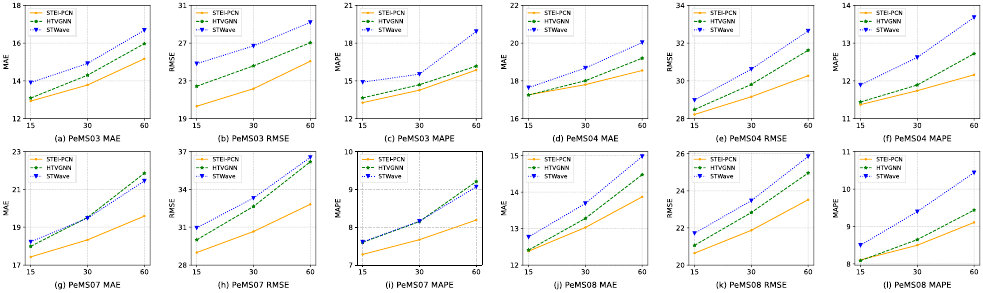}
    \caption{Prediction performance comparison under different prediction horizons.}
    \label{fig:horizon}
\end{figure*}

\subsection{Overall Performance}

\subsubsection{Traffic Flow Prediction}

Table~\ref{tab:flow_results} reports the prediction results on PeMS03, PeMS04, PeMS07, and PeMS08 with prediction horizon $T_p=6$. STEI-PCN achieves the best or second-best results on most metrics across the four flow datasets. Specifically, it obtains the lowest MAE on PeMS03, PeMS07, and PeMS08, and achieves the same best MAE as STD-MAE on PeMS04. In terms of RMSE, STEI-PCN outperforms all baselines on all four datasets, indicating that it effectively suppresses large prediction deviations. For MAPE, STEI-PCN achieves the best results on PeMS04, PeMS07, and PeMS08, and ranks second on PeMS03.

The results verify that STEI-PCN achieves stable prediction performance on road networks of different scales. The performance gains benefit from the integration of local joint spatio-temporal modeling and long-range temporal dependency modeling. The local graph convolution captures propagation-related neighborhood interactions, while TDCN complements longer temporal patterns such as periodicity and trends.

In Fig.~\ref{fig:horizon}, we compare STEI-PCN with several representative advanced baselines under different prediction horizons on PeMS03, PeMS04, PeMS07, and PeMS08. As $T_p$ increases, the prediction horizon becomes longer, and the task becomes more difficult, resulting in increasing MAE, RMSE, and MAPE. STEI-PCN maintains competitive performance across different horizons, especially in short-term prediction.

\subsubsection{Traffic Speed Prediction}

Table~\ref{tab:speed_results} reports the results on PeMS-Bay under different prediction horizons. For short-term prediction with $T_p=3$, STEI-PCN achieves the best MAE, RMSE, and MAPE, showing its strong ability to model short-term speed changes. For medium-term prediction with $T_p=6$, STEI-PCN obtains the best MAE and MAPE, and the second-best RMSE. However, when the prediction horizon increases to $T_p=12$, STEI-PCN falls behind T-Graphormer, STD-MAE, STEP, and RGDAN on several metrics.

This result indicates that long-horizon speed prediction remains challenging for STEI-PCN. Compared with traffic flow prediction, speed prediction is more sensitive to abrupt congestion, recovery processes, and non-local disturbances. Although STEI-PCN effectively captures local propagation patterns and long-range temporal dependencies, its current structure still has limitations in modeling uncertain long-term speed evolution.

\subsubsection{Prediction Visualization}

\begin{figure*}[ht]
    \centering
    \includegraphics[width=0.49\textwidth,trim=0bp 96.6bp 200.15bp 0bp,clip]{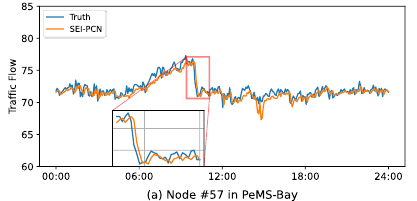}
    \hfill
    \includegraphics[width=0.49\textwidth,trim=200.15bp 96.6bp 0bp 0bp,clip]{Figure/zhenshi.pdf}\\[0.3em]
    \includegraphics[width=0.49\textwidth,trim=0bp 0bp 200.15bp 96.6bp,clip]{Figure/zhenshi.pdf}
    \hfill
    \includegraphics[width=0.49\textwidth,trim=200.15bp 0bp 0bp 96.6bp,clip]{Figure/zhenshi.pdf}
    \caption{Comparison between predicted and ground-truth traffic states.}
    \label{fig:prediction_visualization}
\end{figure*}

Fig.~\ref{fig:prediction_visualization} visualizes the predicted and ground-truth traffic states of randomly selected nodes on PeMS03 and PeMS-Bay over one day with 288 timestamps. The local interval from 5:30 to 6:30 is further enlarged to examine short-term variations. The results show that STEI-PCN can generally track the ground-truth trend and capture abrupt local changes, indicating its ability to model short-term traffic fluctuations.

\subsection{Ablation Study}

\subsubsection{Component Ablation}

To evaluate the contribution of each component, we conduct ablation studies on PeMS04 and PeMS08 with prediction horizon $T_p=12$. This setting is more challenging than the half-hour prediction task with $T_p=6$, and therefore better reflects the role of each module in long-horizon prediction.

The evaluated variants are as follows. ``w/o SCE'' removes the absolute spatial coordinate encoding $Z^S$; ``w/o TCE'' removes the absolute temporal coordinate encoding $Z^T$; ``w/o SDE'' removes the relative spatial distance encoding $Z^{SD}$; and ``w/o TDE'' removes the relative temporal distance encoding $Z^{TD}$. ``w/o STEI'' removes the spatio-temporal encoding and relation inference component and replaces it with an adaptive joint spatio-temporal adjacency matrix. ``w/o STPGAU'' removes the spatio-temporal position-aware gated activation unit. ``w/o GCN'', ``w/o TDCN'', and ``w/o MVC'' remove the graph convolution module, temporal dilated causal convolution module, and multi-view collaborative prediction module, respectively.

As shown in Table~\ref{tab:ablation_components}, the complete STEI-PCN achieves the lowest errors on both datasets. Removing any major component leads to performance degradation, indicating that the performance gain is not caused by a single technique but by the collaboration among spatio-temporal encoding, local graph convolution, long-range temporal convolution, and multi-view prediction.

Among the four encoding types, removing the absolute spatial coordinate encoding causes the most significant degradation. This suggests that node identity plays a fundamental role in dynamic joint spatio-temporal relation inference. Removing STEI or STPGAU also increases the prediction errors, demonstrating the necessity of explicit spatio-temporal relation inference and position-aware feature recalibration. The most severe degradation is observed when GCN or MVC is removed, indicating that local joint spatio-temporal aggregation and multi-view feature fusion are two key structures of STEI-PCN. Removing TDCN also worsens the performance, but the degradation is smaller than that caused by removing GCN or MVC, suggesting that TDCN mainly provides complementary long-range temporal information.

\begin{table*}[ht]
\centering
\scriptsize
\caption{Component ablation results on PeMS04 and PeMS08 with prediction horizon $T_p=12$.}
\label{tab:ablation_components}
\begin{tabular}{lcccccc} 
\toprule
\multirow{3}{*}{Model} & \multicolumn{3}{c}{PeMS04} & \multicolumn{3}{c}{PeMS08} \\
\cmidrule(r){2-4} \cmidrule(r){5-7}
& MAE & RMSE & MAPE & MAE & RMSE & MAPE \\
\midrule
\textbf{STEI-PCN~(Ours)}        & \textbf{18.55} & \textbf{30.27} & \textbf{12.16\%} & \textbf{13.87} & \textbf{23.52} & \textbf{9.11\%} \\
w/o SCE $Z^S$   & 20.75 & 33.32 & 14.04\% & 15.51 & 26.01 & 9.78\% \\
w/o TCE $Z^T$   & 18.63 & 30.33 & 12.19\% & 13.89 & 23.60 & 9.16\% \\
w/o SDE $Z^{SD}$& 18.60 & 30.29 & 12.21\% & 13.91 & 23.67 & 9.13\% \\
w/o TDE $Z^{TD}$& 18.59 & 30.31 & 12.23\% & 13.90 & 23.62 & 9.14\% \\
w/o STEI        & 18.95 & 30.73 & 12.45\% & 14.71 & 24.37 & 9.50\% \\
w/o STPGAU      & 19.45 & 31.44 & 12.78\% & 14.27 & 23.68 & 9.45\% \\
w/o GCN         & 25.80 & 39.99 & 17.42\% & 20.83 & 32.26 & 13.00\% \\
w/o TDCN        & 18.89 & 30.73 & 12.58\% & 14.46 & 23.88 & 9.34\% \\
w/o MVC         & 24.50 & 40.60 & 16.22\% & 22.08 & 40.01 & 13.85\% \\
\bottomrule
\end{tabular}
\end{table*}
\subsubsection{Convergence Analysis}

\begin{figure}[ht]
    \centering
    \includegraphics[width=\linewidth]{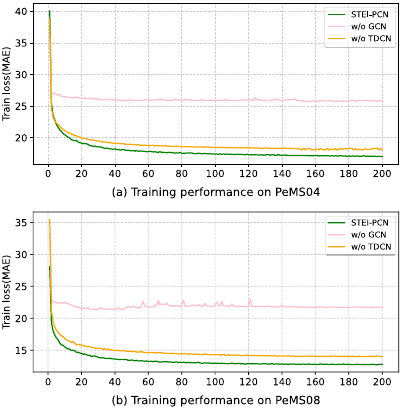}
    \caption{Convergence comparison between STEI-PCN and major ablation variants.}
    \label{fig:convergence}
\end{figure}

Fig.~\ref{fig:convergence} compares the training convergence of STEI-PCN and two major variants, i.e., w/o GCN and w/o TDCN. The complete model achieves lower training error and more stable convergence, which further verifies that the local causal joint spatio-temporal graph and long-range temporal modeling jointly improve the learning process.

\begin{table*}[ht]
\centering
\scriptsize
\caption{Targeted ablation of TPC and TVF on PeMS-Bay with $q=80\%$. Full-sample errors are averaged over three random seeds.}
\label{tab:tpc_tvf}
\begin{tabular}{clcccccccc}
\toprule
 \multirow{3}{*}{$T_p$} & \multirow{3}{*}{Model} & \multicolumn{3}{c}{Full-Sample Error} & \multicolumn{3}{c}{Sharp-Fluctuation Error} & \multicolumn{2}{c}{Sharp-Fluctuation Variation} \\
\cmidrule(r){3-5} \cmidrule(r){6-8} \cmidrule(r){9-10}
 &  & MAE & RMSE & MAPE & MAE & RMSE & MAPE & DirAcc & VarMAE \\
\midrule
\multirow{4}{*}{3}
& STEI-PCN & 1.17 & 2.46 & 2.33\% & 2.9428 & 4.8996 & 6.5297\% & 62.3224 & 2.6426 \\
& STEI-PCN+TPC & 1.15 & 2.40 & 2.30\% & 2.9566 & 4.9473 & 6.4838\% & 64.1898 & 2.6527 \\
& STEI-PCN+TVF & 1.13 & 2.35 & 2.27\% & 2.9561 & 4.8956 & 6.5563\% & 62.6453 & 2.6456 \\
& STEI-PCN+TPC+TVF & 1.14 & 2.38 & 2.30\% & 2.9352 & 4.8991 & 6.4836\% & 63.7360 & 2.6371 \\
\midrule
\multirow{4}{*}{6}
& STEI-PCN & 1.47 & 3.35 & 3.12\% & 3.1971 & 5.6730 & 7.5582\% & 62.7227 & 2.6326 \\
& STEI-PCN+TPC & 1.39 & 3.10 & 2.96\% & 3.5173 & 6.2501 & 8.2783\% & 59.7745 & 2.7194 \\
& STEI-PCN+TVF & 1.40 & 3.13 & 2.96\% & 3.1907 & 5.6745 & 7.4619\% & 62.6679 & 2.6292 \\
& STEI-PCN+TPC+TVF & 1.37 & 3.04 & 2.90\% & 3.5262 & 6.2943 & 8.2897\% & 60.0828 & 2.7271 \\
\midrule
\multirow{4}{*}{12}
& STEI-PCN & 1.88 & 4.42 & 4.28\% & 3.6914 & 6.7477 & 9.0612\% & 60.1583 & 2.7033 \\
& STEI-PCN+TPC & 1.62 & 3.71 & 3.65\% & 3.6714 & 6.6669 & 9.0428\% & 60.3434 & 2.6969 \\
& STEI-PCN+TVF & 1.61 & 3.72 & 3.64\% & 3.6617 & 6.6580 & 9.0352\% & 60.1594 & 2.6941 \\
& STEI-PCN+TPC+TVF & 1.62 & 3.72 & 3.60\% & 3.6866 & 6.7180 & 9.0494\% & 60.1543 & 2.6986 \\
\bottomrule
\end{tabular}
\end{table*}

\begin{table*}[ht]
\centering
\scriptsize
\caption{Threshold sensitivity analysis on PeMS-Bay under sharp speed fluctuations.}
\label{tab:threshold_sensitivity}
\begin{tabular}{cclccc}
\toprule
$T_p$ & Threshold & Model & Sharp MAE & DirAcc & VarMAE \\
\midrule
3 & $q=80\%$ (top 20\%) & STEI-PCN & 2.9428 & 62.3224 & 2.6426 \\
3 & $q=80\%$ (top 20\%) & STEI-PCN+TPC+TVF & 2.9352 & 63.7360 & 2.6371 \\
3 & $q=90\%$ (top 10\%) & STEI-PCN & 4.2613 & 63.6539 & 3.7030 \\
3 & $q=90\%$ (top 10\%) & STEI-PCN+TPC+TVF & 4.2582 & 65.1451 & 3.6941 \\
\midrule
6 & $q=80\%$ (top 20\%) & STEI-PCN & 3.1971 & 62.7227 & 2.6326 \\
6 & $q=80\%$ (top 20\%) & STEI-PCN+TPC+TVF & 3.5262 & 60.0828 & 2.7271 \\
6 & $q=90\%$ (top 10\%) & STEI-PCN & 4.5753 & 65.3610 & 3.6566 \\
6 & $q=90\%$ (top 10\%) & STEI-PCN+TPC+TVF & 5.1132 & 62.0628 & 3.8223 \\
\midrule
12 & $q=80\%$ (top 20\%) & STEI-PCN & 3.6914 & 60.1583 & 2.7033 \\
12 & $q=80\%$ (top 20\%) & STEI-PCN+TPC+TVF & 3.6866 & 60.1543 & 2.6986 \\
12 & $q=90\%$ (top 10\%) & STEI-PCN & 5.1992 & 62.7973 & 3.7618 \\
12 & $q=90\%$ (top 10\%) & STEI-PCN+TPC+TVF & 5.1844 & 62.8668 & 3.7543 \\
\bottomrule
\end{tabular}
\end{table*}

\begin{table*}[ht]
\centering
\scriptsize
\caption{Parameter size and runtime comparison on PeMS04 and PeMS07. Training and inference time are reported in seconds per epoch.}
\label{tab:cost}
\begin{tabular}{lcccccc}
\toprule
\multirow{2}{*}{Model} & \multicolumn{3}{c}{PeMS04} & \multicolumn{3}{c}{PeMS07} \\
\cmidrule(r){2-4} \cmidrule(r){5-7}
 & Params & Training & Inference & Params & Training & Inference \\
\midrule
DSTAGNN (2022) & 3.58M & 56.74 & 3.16 & 14.41M & 628.65 & 36.55 \\
STPGCN (2022) & 0.20M & 35.39 & 1.12 & 0.20M & 333.70 & 15.98 \\
DGCRN (2023) & 0.22M & 21.50 & 1.16 & 0.22M & 138.88 & 7.72 \\
PDFormer (2023) & 0.53M & 18.37 & 1.59 & 0.53M & 154.31 & 4.97 \\
SSGCRTN (2024) & 0.38M & 29.23 & 1.44 & 0.46M & 86.76 & 3.18 \\
\textbf{STEI-PCN~(Ours)} & 0.45M & 10.94 & 0.25 & 0.46M & 107.21 & 1.42 \\
\bottomrule
\end{tabular}
\end{table*}

\subsection{Analysis of TPC and TVF}

\subsubsection{Targeted Ablation on PeMS-Bay}

TPC and TVF are designed as training-stage auxiliary constraints for speed prediction under sharp speed changes. Therefore, their effectiveness is further evaluated on PeMS-Bay. We compare four training versions: the basic STEI-PCN trained only with the prediction loss, STEI-PCN+TPC, STEI-PCN+TVF, and STEI-PCN+TPC+TVF. All versions use the same data split, historical input length, and backbone structure. The input length is fixed to $T_h=12$, and the prediction horizon is set to $T_p\in\{3,6,12\}$.

Table~\ref{tab:tpc_tvf} reports the full-sample prediction errors and the results on sharp speed-change samples. The full-sample MAE, RMSE, and MAPE are averaged over three random seeds. The sharp-fluctuation metrics are computed from representative complete evaluation results and are used to analyze the behavior of different constraints on abrupt speed-change segments. TPC and TVF are only added to the training objective, and no ground-truth future states or additional online computation branches are introduced during inference.

\subsubsection{Sharp-Fluctuation Evaluation}

To evaluate the model under speed-change scenarios, we select sharp speed-change samples based on the first-order variation magnitude of the ground-truth future speed sequence. For node $v_i$ and prediction step $k$, the variation magnitude is defined as
\begin{equation}
a_i^k=\left|\Delta X^P(k,i)\right|.
\end{equation}
A larger $a_i^k$ indicates a stronger speed variation. We select the samples whose variation magnitudes belong to the top $20\%$ in the test set. The sharp-fluctuation sample set is denoted as
\begin{equation}
\mathcal{H}_q=
\{(k,i)\mid a_i^k\geq Q_q(\{a_j^r\})\},
\end{equation}
where $Q_q(\cdot)$ denotes the $q$-quantile threshold. In Table~\ref{tab:tpc_tvf}, $q=80\%$ is used, corresponding to the top $20\%$ largest speed variations.

For $T_p=3$, all enhanced versions reduce full-sample errors compared with the basic STEI-PCN. STEI-PCN+TVF achieves the lowest full-sample MAE, RMSE, and MAPE. On sharp speed-change samples, TPC improves DirAcc, indicating that propagation consistency helps short-term speed-change direction prediction. The combination of TPC and TVF slightly reduces sharp-fluctuation MAE, MAPE, and VarMAE.

For $T_p=6$, the enhanced versions still improve full-sample prediction errors. However, the variants containing TPC perform worse on sharp speed-change samples, while TVF alone shows more stable behavior and slightly reduces sharp-fluctuation MAE, MAPE, and VarMAE. This suggests that local propagation consistency may conflict with medium-term non-local or periodic changes in some speed-change scenarios.
the 
For $T_p=12$, all enhanced versions reduce full-sample errors. On sharp speed-change samples, both TPC and TVF improve MAE, RMSE, MAPE, and VarMAE compared with the basic model. TVF obtains the best sharp-fluctuation error and VarMAE, while TPC slightly improves DirAcc. However, the combination of TPC and TVF does not always outperform TVF alone, indicating that the two constraints do not provide simply additive gains.

Overall, TPC and TVF should be understood as targeted training mechanisms for speed prediction under sharp speed changes. Their effects depend on prediction horizon and sample type, rather than guaranteeing uniform improvement across all metrics and all settings.
\subsubsection{Threshold Sensitivity}

To further examine whether the conclusion is sensitive to the definition of sharp speed-change samples, we conduct a lightweight threshold sensitivity analysis on PeMS-Bay. In addition to $q=80\%$, we use a stricter threshold $q=90\%$, corresponding to the top $10\%$ largest speed variations.

As shown in Table~\ref{tab:threshold_sensitivity}, STEI-PCN+TPC+TVF reduces sharp-fluctuation MAE and VarMAE under both thresholds for $T_p=3$ and $T_p=12$, and improves DirAcc in most cases. However, for $T_p=6$, the complete constrained version performs worse than the basic STEI-PCN under both thresholds, which is consistent with the observation in Table~\ref{tab:tpc_tvf}. Therefore, the effects of TPC and TVF are horizon-dependent. TVF provides relatively stable control of variation magnitude, while TPC is more helpful for short-term direction consistency. Their combination can benefit short-term and some long-horizon sharp speed-change cases, but it does not guarantee monotonic improvement under all settings.

\subsection{Hyperparameter Sensitivity}

\begin{figure}[!t]
    \centering
    \includegraphics[width=\linewidth]{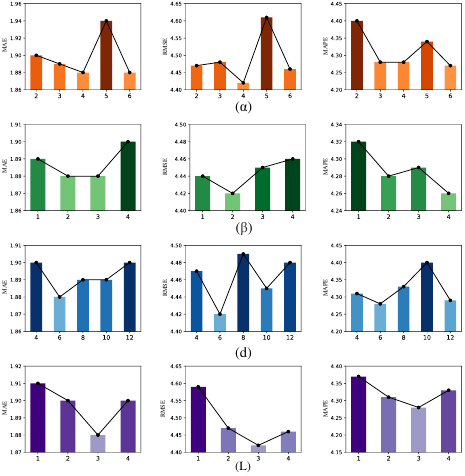}
    \caption{Influence of key hyperparameters on prediction performance.}
    \label{fig:hyperparameter}
\end{figure}

We further analyze the influence of key hyperparameters, including the spatial interaction range $\alpha$, temporal interaction range $\beta$, encoding dimension $d$, and the number of TDCN layers $L$. When one hyperparameter is varied, the others are fixed to their default values. We present related results of PeMS-Bay in Fig.~\ref{fig:hyperparameter}.

The model performance first improves and then degrades as $\alpha$ or $\beta$ increases. This indicates that an appropriate spatio-temporal interaction range helps balance local neighborhood information and broader contextual information. A small interaction range limits the ability to capture local synchronous joint spatio-temporal dependencies, while an excessively large range may introduce weakly related nodes or noisy historical states.

For the encoding dimension $d$, a relatively small dimension is sufficient to represent spatio-temporal positions and distances. Increasing $d$ does not necessarily lead to better performance, but may introduce redundant parameters. For the number of TDCN layers, STEI-PCN achieves the best performance with three layers. With kernel size $3$ and dilation factors $(1,2,4)$, the effective receptive field is sufficient to cover the historical input window used in this paper.

The coefficients of TPC and TVF are controlled by $\lambda$ and $\gamma$, respectively, and the fluctuation weight in TVF is controlled by $\eta$. Since TPC and TVF are positioned as auxiliary training mechanisms for sharp speed fluctuations rather than backbone hyperparameters, we do not conduct a large-scale coefficient sensitivity study in this paper. In the targeted ablation experiments, we set $\lambda=0.01$, $\gamma=0.02$, and $\eta=1.0$. Further coefficient sensitivity and cross-dataset validation are left for future work.

\subsection{Computational Cost}
Table~\ref{tab:cost} compares the parameter size, training time, and inference time of STEI-PCN with several advanced baselines on PeMS04 and PeMS07. STEI-PCN keeps the parameter size at a moderate level, i.e., $0.45$M on PeMS04 and $0.46$M on PeMS07. Compared with DSTAGNN, STEI-PCN uses far fewer parameters. Compared with STPGCN, STEI-PCN achieves lower inference time, especially on the large-scale PeMS07 dataset.

On PeMS04, STEI-PCN achieves the lowest training and inference time among the compared models. On PeMS07, its training time is slightly higher than SSGCRTN, but its inference time is lower. These results show that the pure convolutional architecture and the single-layer local joint spatio-temporal graph convolution reduce the online prediction cost while maintaining competitive prediction accuracy.

TPC and TVF do not introduce new trainable parameters and do not change the forward inference structure. TPC computes propagation states on the local spatio-temporal edge support only during training, and TVF computes first-order differences from the prediction outputs. Therefore, their additional costs are limited to the training stage and do not affect online inference.

\section{Conclusion and Future Work}
\label{sec:conclusion}

This work presents STEI-PCN, an efficient pure convolutional network for traffic prediction based on spatio-temporal encoding and relation inference. The proposed STEI module calculates dynamic edge weights for local joint spatio-temporal graphs using position and distance encodings, and single-layer graph convolution is adopted to aggregate local propagation information.

To improve speed prediction under sharp speed-change scenarios, this paper further introduced two training-stage auxiliary constraints, namely traffic propagation consistency (TPC) and traffic variation fidelity (TVF). TPC encourages the hidden representation of each target node to be consistent with its locally propagated neighborhood state, while TVF constrains the first-order variations of the predicted sequence to preserve the direction and magnitude of speed changes. These two constraints do not change the inference structure or introduce additional online prediction cost.

Experiments on four traffic flow datasets and one traffic speed dataset show that STEI-PCN achieves competitive prediction accuracy with a moderate parameter size and low computational cost. Ablation studies verify the effectiveness of the main components, including STEI, local joint spatio-temporal graph convolution, TDCN, and multi-view prediction. The targeted experiments on PeMS-Bay indicate that TPC and TVF can provide complementary benefits for some sharp speed-fluctuation cases, but their effects depend on prediction horizon and sample type. In particular, STEI-PCN still has limitations in long-horizon speed prediction on PeMS-Bay, suggesting that modeling uncertain long-range speed evolution remains challenging.

Future work will focus on improving the robustness and generalization ability of STEI-PCN. First, more informative spatio-temporal encodings can be designed by incorporating road hierarchy, region attributes, travel demand, and other external factors. Second, multi-source information such as weather, incidents, holidays, and traffic control signals can be integrated to better distinguish periodic patterns and abnormal disturbances. Third, uncertainty estimation, online updating, cross-city transfer, and model compression can be further explored to support practical deployment. Finally, the proposed spatio-temporal encoding and relation inference framework may be extended to other graph-based spatio-temporal forecasting tasks.

\bibliographystyle{IEEEtran}
\bibliography{ref}
\end{document}